\documentclass[11pt,a4paper]{article}
\usepackage[hyperref]{emnlp2018}
\usepackage{times}
\usepackage{latexsym}

\usepackage[utf8]{inputenc} 
\usepackage{xcolor,colortbl}
\usepackage{graphicx}
\usepackage{url}
\usepackage{bbold}
\usepackage{makecell}
\usepackage[titletoc,title]{appendix}
\usepackage{enumitem}
\usepackage{float}
\usepackage{pgfplots}

\aclfinalcopy 

\definecolor{Gray}{gray}{0.90}
\definecolor{LightCyan}{rgb}{0.88,1,1}

\newcolumntype{a}{>{\columncolor{Gray}}c}
\newcolumntype{b}{>{\columncolor{white}}c}

\usepackage[normalem]{ulem}
\definecolor{darkgreen}{rgb}{0.0, 0.4, 0.0}

\def\ODdel#1{\bgroup\markoverwith{\textcolor{darkgreen}{\rule[0.5ex]{2pt}{1pt}}}\ULon{#1}}

\title{Neural Response Ranking for Social Conversation:\\ A Data-Efficient Approach}

\author{Igor Shalyminov, Ondřej Dušek, and Oliver Lemon \\
  The Interaction Lab, Department of Computer Science\\
  Heriot-Watt University, 
  Edinburgh, EH14 4AS, UK\\
  {\tt \{is33, o.dusek, o.lemon\}@hw.ac.uk}\\
  }

\date{}

\begin{document}
\maketitle
\begin{abstract}
 The overall objective of `social' dialogue systems is to support engaging, entertaining, and lengthy conversations on a wide variety of topics, including social chit-chat.
Apart from raw dialogue data, user-provided ratings are the most common signal used to train such systems to produce engaging responses.
In this paper we show that social dialogue systems can be trained effectively from raw unannotated data. Using a dataset of real conversations collected in the 2017 Alexa Prize challenge, we developed a neural ranker\footnote{Code and trained models are available at \url{https://github.com/WattSocialBot/alana_learning_to_rank}} for selecting `good' system responses to user utterances, i.e.\ responses which are likely to lead to long and engaging conversations. We show that (1) our neural ranker consistently outperforms several strong baselines when trained to optimise for user ratings; (2) when trained on larger amounts of data and only using conversation length as the objective, the ranker performs better than the one trained using ratings~-- ultimately reaching a Precision@1 of $0.87$.
This advance will make data collection for social conversational agents simpler and less expensive in the future.
\end{abstract}

\section{Introduction}
Chatbots, or \textit{socialbots}, are dialogue systems aimed at maintaining an open-domain conversation with the user spanning a wide range of topics, with the main objective  of being engaging, entertaining, and natural. Under one of the current approaches to such systems, the {\it bot ensemble} \cite{mila,yu_strategy_2016,gen_ir_ensemble}, a collection, or ensemble, of different bots is used, each of which proposes a candidate response to the user's input, and a \emph{response ranker}  selects the best response for the final system output to be uttered to the user.

In this paper, we focus on the task of finding the best supervision signal for training a response ranker for ensemble systems.
Our contribution is twofold:
first, we present a neural ranker for ensemble-based dialogue systems and evaluate its level of performance using an annotation type which is often used in open-domain dialogue and was provided to the Alexa Prize 2017 participants by Amazon \cite{ram_conversational_2017}: per-dialogue user ratings.
Second and most importantly, we explore an alternative way of assessing social conversations simply via their {\it length}, thus removing the need for any user-provided ratings. 


\section{Data Efficiency in Social Dialogue}

\subsection{The Need for Data Efficiency}

It is well known that deep learning models are highly data-dependent, but there are  currently no openly available data sources which can provide enough high-quality open-domain social dialogues for building a production-level socialbot. Therefore, a common way to get the necessary data is to collect it on a crowdsourcing platform \cite{edina}. Based on the model type and the development stage, it may be necessary to collect either whole dialogues, or some form of human feedback on how good a particular dialogue or turn is. However, both kinds of data are time-consuming and expensive to collect.

The data efficiency of a dialogue model can be split into two parts accordingly:
\begin{itemize}[itemsep=-4pt,topsep=0pt,leftmargin=10pt,labelwidth=10pt]
\item \textit{sample efficiency}~-- the number of data points needed for the model to train. As such, it is useful to specify an order of magnitude of the training set size for different types of machine learning models;
\item \textit{annotation efficiency}~-- the amount of annotation effort needed. For instance, traditional goal-oriented dialogue system architectures normally require \textit{intent}, \textit{slot value}, and \textit{dialogue state} annotation \citep[e.g.][]{young_hidden_2010}, whereas end-to-end conversational models work simply with raw text transcriptions \citep[e.g.][]{DBLP:journals/corr/VinyalsL15}.
\end{itemize}

\subsection{Alexa Prize Ratings}

\begin{table}[t]
\begin{center}
\small
\begin{tabular}{|l|c|}\hline
\bf Variables& \bf Pearson corr. coefficient\\\hline\hline
rating/length&$0.11$\\
\rowcolor{Gray}
rating/positive feedback&$0.11$\\
rating/negative feedback&$0.04$\\
\rowcolor{Gray}
length/positive feedback&$0.67$\\
length/negative feedback&$0.49$\\\hline
\end{tabular}
\end{center}
\caption{Correlation study of key dialogue aspects}\label{tab:correlation}
\end{table}

The 2017 Alexa Prize challenge made it possible to collect large numbers of dialogues between real users of Amazon Echo devices and various chatbots. The only annotation collected was per-dialogue ratings elicited at the end of conversations by asking the user {\it ``On a scale of 1 to 5, how much would you like to speak with this bot again''}  \cite{venkatesh_evaluating_2017}. Less than 50\% of conversations were actually rated; the rest were quit without the user giving a score. In addition, note that a single rating is applied to an entire conversation (rather than individual turns), which may consist of very many utterances. The conversations in the challenge were about 2.5 minutes long on average, and about 10\% of conversations  were over 10 minutes long \cite{ram_conversational_2017} -- this makes the ratings very sparse. Finally, the ratings are noisy -- some dialogues which are clearly bad can get good ratings from some users, and vice-versa.

Given the main objective of social dialogue stated in the Alexa Prize rules as `long and engaging' conversation, we tried to verify an assumption that user ratings reflect these properties of the dialogue. Apart from our observations above, we performed a correlation analysis of user ratings and aspects of dialogue directly reflecting the objective: dialogue length and explicit user feedback (see Table \ref{tab:correlation}).

Although we have a significant number of dialogues which are both long and highly rated, the correlation analysis was not able to show any relationship between dialogue length and rating. Neither are ratings correlated with user feedback (see Section~\ref{sec:eval} for the details of user feedback collection).
On the other hand, we found a promising moderate correlation between the conversation length and explicit positive feedback from users (specifically, the number of dialogue turns containing it). The respective length/negative feedback relationship is slightly weaker.

Therefore, we experiment with conversation length for approximating user satisfaction and engagement and use it as an alternative measure of dialogue quality.
This allows us to take advantage of all conversations, not just those rated by users, for training a ranker.
While some conversations might be long but not engaging (e.g.\ if there are a lot of misunderstandings, corrections, and speech recognition errors), training a ranker only using length makes it extremely {\it annotation-efficient}. 

\section{A neural ranker for open-domain conversation}
\label{sec:neural}

\begin{figure}
\begin{center}
\includegraphics[width=0.95\columnwidth]{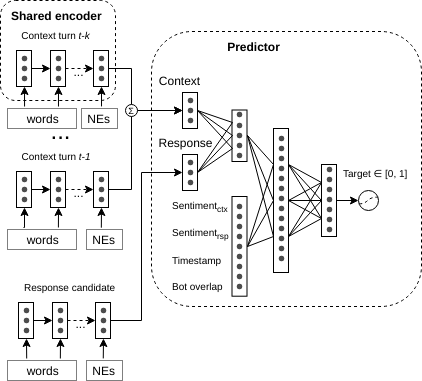}
\end{center}
\caption{Neural ranker architecture}\label{fig:ranker_architecture}
\end{figure}


The ranker described here is part of Alana, Heriot-Watt University's Alexa Prize 2017 finalist socialbot \cite{AlanaNIPS}. Alana is an ensemble-based model incorporating information-retrieval-based bots with news content and information on a wide range of topics from Wikipedia, a question answering system, and rule-based bots for various purposes, from amusing users with fun facts to providing a consistent persona. The rule-based bots are also required to handle sensitive issues which can be raised by real users, such as medical, financial, and legal advice, as well as profanities. 

\subsection{Ranker architecture}\label{sec:architecture}

The architecture of our ranker is shown in Figure \ref{fig:ranker_architecture}.
The inputs to the model are 1-hot vectors of a candidate response and the current dialogue context (we use the 3 most recent system and user turns). They are encoded into a latent representation using a single shared RNN encoder based 
on GRU cells \cite{cho-al-emnlp14}. The context embedding vectors are then summed up and concatenated with the response embedding (Eq.~\ref{eq:encoder}):

\begin{equation}
\label{eq:encoder}
Enc(C, r) = \sum_{i}{\textit{RNN}(C_i)} \oplus \textit{RNN}(r)
\end{equation}
where $C$ is the dialogue context and $r$ is a response candidate.

The context and the response are represented using combined word-agent tokens (where agent is either a specific bot from the ensemble or the user) and  are concatenated with the lists of named entities extracted using Stanford NER \cite{StanfordNER}. All the word-agent tokens and named entities share the same unified vocabulary.

Encoder outputs, along with additional dialogue features such as context and response sentiment, timestamp, and bot names in the context and the response, go into the \textit{Predictor}, a feed-forward neural network (MLP) whose output is the resulting rating (Eq.~\ref{eq:predictor}):

\begin{equation}
\label{eq:predictor}
Pred(C, r) = \sigma(L(Sem(C, r) \oplus f(C, r)))
\end{equation}
\begin{itemize}[itemsep=-4pt,topsep=0pt,leftmargin=3.5em,labelwidth=*,align=left]
  \item[where:] $L(x) = \textit{ReLU}(M x + b)$ is the 
  layer used in the Predictor (the number of such layers is a model parameter),
  \item[] $Sem = L(Enc(C, r))$ is the vector of semantic context-response features, and
  \item[] $f(C, r)$ is a vector of the additional dialogue features listed above.
\end{itemize}
\vspace{2mm}
We use $ReLU$ activation for the hidden layers because it is  known to be highly efficient with deep architectures \cite{relu}. Finally, we use sigmoid activation $\sigma$ for generating the final prediction in the range $[0, 1]$.

\subsection{Training method}\label{sec:train-method}

We use either dialogue rating or length as the prediction target (as discussed in Sections~\ref{sec:dataset} and~\ref{sec:eval}). 
The model is trained to minimize the Mean Squared Error (MSE) loss against the target using the Adagrad optimizer \cite{adagrad}.
In our training setup, the model learns to predict per-turn target values. However, since only per-dialogue ones are available in the data, we use the following approximation: the target value of a context-response pair is the target value of the dialogue containing it. The intuition behind this is an assumption that the majority of turns in ``good" dialogues (either length- or rating-wise) are ``good" in their local contexts as well~-- so that given a large number of dialogues, the most successful and unsuccessful turns will emerge from the corresponding dialogues.


\section{Baselines}\label{sec:baselines}

We compare our neural ranker to two other models also developed during the competition: \textit{handcrafted} and \textit{linear} rankers~--- all three were deployed live in the Alana Alexa Prize 2017 finalist system \cite{AlanaNIPS}, and were therefore of sufficient quality for a production system receiving thousands of calls per day. We also compare our model to a recently published \textit{dual-encoder} response selection model by \citet{lu_practical_2017} based on an approach principally close to ours.

\subsection{Handcrafted ranker}
In the handcrafted   approach, several turn-level and dialogue-level features are calculated, and a linear combination of those feature values with manually adjusted coefficients is  used to predict the final ranking. The list of features includes:

\begin{itemize}[itemsep=-4pt,topsep=0pt,leftmargin=10pt,labelwidth=10pt]
\item coherence, information flow, and dullness as defined by \citet{Li2016};
\item overlap between the context and the response with regards to named entities and noun phrases;
\item topic divergence between the context turns and the response -- topics are represented using the \textit{Latent Dirichlet Allocation} (LDA) model \cite{lda};
\item sentiment polarity, as computed by the NLTK Vader sentiment analyser \cite{gilbert_vader:_2014}.\footnote{\url{http://www.nltk.org/howto/sentiment.html}}
\end{itemize}

\subsection{Linear ranker}\label{sec:linear-ranker}

The linear ranker is based on the VowpalWabbit (VW) linear model \cite{vowpalwabbit}. 
We use the MSE loss function and the following features in our VW ranker model:
\begin{itemize}[itemsep=-4pt,topsep=0pt,leftmargin=10pt,labelwidth=10pt]
\item bag-of-n-grams from the dialogue context (preceding 3 utterances) and the response,
\item position-specific n-grams at the beginning of the context and the response (first 5 positions),
\item dialogue flow features \cite{Li2016}, the same as for the handcrafted ranker,
\item bot name, from the set of bots in the ensemble.
\end{itemize}

\subsection{Dual-encoder ranker}
The closest architecture to our neural ranker is that of \cite{lu_practical_2017}, who use a dual-encoder LSTM with a predictor MLP for task-oriented dialogue in closed domains. Unlike this work, they do not use named entities, sentiment, or other input features than basic word embeddings. Dialogue context is not modelled explicitly either, and is limited to a single user turn. We reproduced their architecture 
and set its parameters to the best ones reported in the original paper.

\section{Training data}\label{sec:dataset}


\noindent
Our data is transcripts of conversations between our socialbot and real users of the Amazon Echo collected over the challenge period, February--December 2017.
The dataset consists of 
over 200,000 dialogues (5,000,000+ turns) from which 
over 100,000 dialogues (totalling nearly 3,000,000 turns) are annotated with ratings.
From this data, we sampled two datasets of matching size for training our rankers, using the per-turn target value approximation described in Section~\ref{sec:train-method} -- the \emph{Length} and \emph{Rating} datasets for the respective versions of rankers.


The target values (length/rating) in both sets are normalized into the $[0,1]$ range, and the \emph{Length} set contains context-response pairs from long dialogues (target value above $0.7$) as positive instances and context-response pairs from short dialogues (target value below $0.3$) as negative ones.
With the same selection criteria, the \emph{Rating} set contains context-response pairs from highly rated dialogues (ratings 4 and 5) as positive instances and context-response pairs from low-rated dialogues (ratings 1 and 2) as negative ones.
Both datasets contain 500,000 instances in total, with equal proportion of positive and negative instances. We use a 8:1:1 split for training, development, and  test sets.

Prior to creating both datasets, we filtered out of the dialogue transcripts all system turns which cannot be treated as natural social interaction (e.g.\ a quiz game) as well as outliers (interaction length $\ge 95$th percentile or less than 3 turns long).%
\footnote{Some extremely long dialogues are due to users repeating themselves over and over, and so this filter removes these bad dialogues from the dataset. Dialogues less than 3 turns long are often where the user accidentally triggered the chatbot. These outliers amounted to about 14\% of our data.}
Thresholds of $0.3$ and $0.7$ were set heuristically based on preliminary data analysis. On the one hand, these values provide contrastive-enough ratings (e.g.\ we are not sure whether the rating in the middle of the scale can be interpreted as negative or positive). On the other hand, they allow us to get enough training data for both Length and Rating datasets.\footnote{Using more extreme thresholds did not produce enough data while less ones did not provide adequate training signal.}


\section{Evaluation and experimental setup}\label{sec:eval}
In order to tune the neural rankers, we performed a grid search over the shared encoder GRU layer size and the Predictor topology.\footnote{We tested GRU sizes of 64, 128, 256 and Predictor layers number/sizes of [128], [128, 64], [128, 32, 32].} The best configurations are determined by the loss on the development sets. For evaluation, we used an independent dataset.


\subsection{Evaluation based on explicit user feedback}\label{sec:user-feedback-dataset}

At the evaluation stage, we check how well the rankers can distinguish between good responses and bad ones. The criterion for `goodness' that we use here is chosen to be independent from both training signals. Specifically, we collected an evaluation set composed of dialogue turns followed by explicit user feedback, e.g.\ ``great, thank you", ``that was interesting" (we refer to it as the \emph{User feedback} dataset). Our `bad' response candidates are randomly sampled across the dataset.

The user feedback turns were identified using sentiment analysis in combination with a whitelist and a blacklist of hand-picked phrases, so that in total we used 605 unique utterances, e.g. \textit{``that's pretty cool'', ``you're funny'', ``gee thanks'', ``interesting fact'', ``funny alexa you're funny''}. 

`Goodness' defined in this way allows us to evaluate how well our two approximated training signals can optimize for the user's satisfaction as explicitly expressed at the turn level, thus leading to our desired behaviour, i.e., producing long and engaging dialogues.

The \emph{User feedback} dataset contains 24,982 $\langle context, good\_response, bad\_response \rangle$ tuples in total.

To evaluate the rankers on this dataset, we use \textit{precision@k}, which is commonly used for information retrieval system evaluation (Eq.~\ref{eq:precision_at_k}).

\begin{equation} \label{eq:precision_at_k}
P@k(c, R) = \frac{\sum_{i=1}^k{Relevant(c, R_k)}}{k}
\end{equation}
where $c$ is dialogue context, $R$ is response candidates list, and $Relevant$ is a binary predicate indicating whether a particular response is relevant to the context.

Precision is typically used together with recall and F-measure. However, since our dialogue data is extremely sparse so that it is hard to find multiple good responses for the same exact dialogue context, recall and F-measure cannot be applied to this setting.
Therefore, since we only perform pairwise ranking, we use \textit{precision@1} to check that the good answer is the top-ranked one.
Also due to data sparsity, we only perform this evaluation with \textit{gold positive} responses and \textit{sampled negative} ones~-- it is typically not possible to find a good response with exactly the same context as a given bad response.


\subsection{Interim results}

The results of our first experiment are shown in Table~~\ref{tab:ranker_eval}. We can see that the neural ranker trained with user ratings clearly outperforms all the alternative approaches in terms of test set loss on its respective dataset as well as pairwise ranking precision on the evaluation dataset.
Also note that both versions of the neural ranker stand extremely close to each other on both evaluation criteria, given a much greater gap between them and their next-best-performing alternatives, the linear rankers.

The dual-encoder ranker turned out to be not an efficient model for our problem, partly because it was originally optimized for a different task as reported by \citet{lu_practical_2017}.


\section{Training on larger amounts of data}\label{sec:larger-data}
\begin{table}[h]
\centering
\small
\begin{tabular}{|l|c|c|c|}\hline
\bf Model&\makecell[c]{\bf P@1\\\bf (eval set)}&\makecell[c]{\bf Loss\\\bf (test set)}\\\hline\hline
Handcrafted&0.478&---\\
\rowcolor{Gray}
VowpalWabbit@length&0.742&0.199\\
VowpalWabbit@rating&0.773&0.202\\
\rowcolor{Gray}
DualEncoder@length&0.365&0.239\\
DualEncoder@rating&0.584&0.247\\
\rowcolor{Gray}
Neural@length&0.824&0.139\\
Neural@rating&\textbf{0.847}&\textbf{0.138} \\ \hline
\end{tabular}
\caption{Ranking models evaluation: pairwise ranking precision on the independent \emph{User feedback} dataset and loss on the \emph{Length/Rating} test sets (Section~\ref{sec:dataset}) for the corresponding trainset sizes of 500,000.}
\label{tab:ranker_eval}
\end{table}

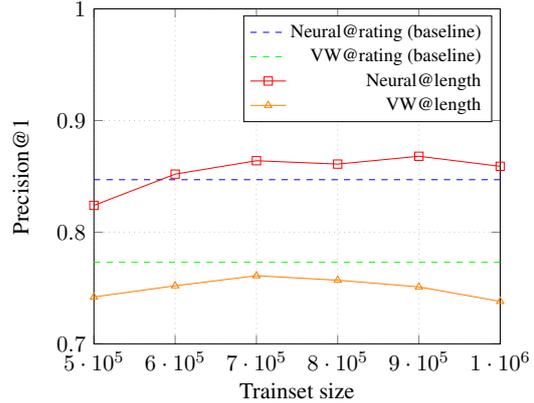
\begin{figure}
\centering
\pgfplotsset{scaled x ticks=false}
\begin{tikzpicture}[thick,scale=0.78, every node/.style={transform shape}]
\begin{axis}[
    xlabel={Trainset size},
    ylabel={Precision@1},
    xmin=500000, xmax=1000000,
    ymin=0.7, ymax=1,
    xtick={500000, 600000, 700000, 800000, 900000, 1000000},
    legend style={font=\small},
    legend cell align={right},
    ymajorgrids=true,
    xmajorgrids=true,
    grid style=dotted,
]
 
\addplot[dashed, color=blue]
coordinates {
    (500000,0.847)(600000,0.847)(700000,0.847)(800000,0.847)(900000,0.847)(1000000,0.847)
};
\addplot[dashed, color=green]
coordinates {
    (500000,0.773)(600000,0.773)(700000,0.773)(800000,0.773)(900000,0.773)(1000000,0.773)
};
\addplot[color=red, mark=square]
coordinates {
    (500000,0.824)(600000,0.852)(700000,0.864)(800000,0.861)(900000,0.868)(1000000,0.859)
};
\addplot[color=orange, mark=triangle]
coordinates {
    (500000,0.742)(600000,0.752)(700000,0.761)(800000,0.757)(900000,0.751)(1000000,0.738)
};
\legend{Neural@rating (baseline), VW@rating (baseline), Neural@length, VW@length}
\end{axis}
\end{tikzpicture}
\caption{Comparison of rankers trained on extended datasets}
\label{fig:extended_datasets}
\end{figure}

A major advantage of training on raw dialogue transcripts is data volume: in our case, we have roughly twice as many raw dialogues as rated ones (cf.~Section~\ref{sec:dataset}). This situation is very common in data-driven development: since data annotation is a very expensive and slow procedure, almost always there is significantly more raw data than annotated data of a high quality.
To illustrate this, we collected extended training datasets of raw dialogues of up to 1,000,000 data points for training from the length signal. We trained our neural ranker and the VW ranker using the same configuration as in Section~\ref{sec:eval}.\footnote{We were not able to train the dual encoder ranker on all the extended datasets due to the time constraints.}

The results are shown in Figure \ref{fig:extended_datasets}, where we see that the neural ranker trained on the length signal consistently outperform the ratings-based one. Its trend, although fluctuating, is more stable than that of VW~-- we believe that this is due to VW's inherent lower model capacity as well as its training setup, which is mainly optimised for speed. The figure also shows that VW@length is worse than VW@rating, regardless of training data size.

\section{Discussion and future work}\label{sec:discussion}

Our evaluation results show that the neural ranker presented above is an efficient approach to response ranking for social conversation. On a medium-sized training set, the two versions of the neural ranker, length and ratings-based, showed strongly superior performance to three alternative ranking approaches, and performed competitively with each other.
Furthermore, the experiment with extended training sets shows that the accuracy of the length-based neural ranker grows steadily given more unannotated training data, outperforming the rating-based ranker with only slightly larger training sets.

The overall results of our experiments confirm that dialogue length, even approximated in quite a straightforward way, provides a sufficient supervision signal for training a ranker for a social conversation model. 
In future work, we will attempt to further improve the model using the same data in an adversarial setup following \citet{irgan}. We also plan to directly train our model for pairwise ranking in the fashion of \citet{ranknet} instead of the current pointwise approach.
Finally, we are going to employ contextual sampling of negative responses using approximate nearest neighbour search \cite{JDH17} in order to perform a more efficient pairwise training.

\section{Related work}\label{sec:related}

Work on response ranking for conversational systems has been been growing rapidly
in recent years.
Some authors employ ranking based on heuristically defined measures:
\citet{yu_ticktock:_2015,yu_strategy_2016} use a heuristic based on keyword matching, part-of-speech filters, and Word2Vec similarity. \cite{edina} apply standard information retrieval metrics (TF-IDF) with importance weighting for named entities.
However, most of the recent research attempts to train the ranking function from large amounts of conversational data, as we do.
Some authors use task-based conversations, such as IT forums \cite{ubuntu_corpus} or customer services \cite{lu_practical_2017,kumar_question-answer_2018}, while
others focus on online conversations on social media \citep[e.g.][]{wu_ranking_2016,al-rfou_conversational_2016}.

The basic approach to learning the ranking function in most  recent work is the same  \citep[e.g.][]{ubuntu_corpus,al-rfou_conversational_2016,wu_ranking_2016}: the predictor is taught to rank positive responses taken from real dialogue data higher than randomly sampled negative examples.
Some of the approaches do not even include rich dialogue contexts and use only immediate context-response pairs for ranking \cite{ji_information_2014,yan_learning_2016,lu_practical_2017}.
Some authors improve upon this basic scenario: \citet{zhuang_ensemble_2018} take a desired emotion of the response into account; \citet{liu_rubystar:_2017} focus on the engagement of responses based on Reddit comments rating; \citet{fedorenko_avoiding_2017} train the ranking model in several iterations, using highly ranked incorrect responses as negative examples for the next iteration.
Nevertheless, to our knowledge, none of the prior works attempt to optimise for long-term dialogue quality; 
unlike in our work, their only ranking criterion is focused on the immediate response.


\section{Conclusion}\label{sec:conclusion}
We have presented a neural response ranker for open-domain `social' dialogue systems and described two methods for training it using common supervision signals coming from conversational data: user-provided ratings and dialogue length. We demonstrated its efficiency by evaluating it using explicit positive feedback as a measure for user engagement. Specifically, trained on ratings, our neural ranker consistently outperforms several strong baselines; moreover, given larger amounts of data and only using conversation length as the objective, the ranker performs better the ratings-based one, reaching $0.87$ Precision@1.
This shows that conversation length can be used as an optimisation objective for generating engaging social dialogues, which means that we no longer need the expensive and time-consuming procedure of collecting per-dialogue user ratings, as was done for example in the Alexa Prize 2017 and is common practice in conversational AI research.
Per-turn user ratings may still be valuable to collect for such systems, but these are even more expensive and problematic to obtain.
Looking ahead, this advance will make data collection for social conversational agents simpler and less expensive in the future.

\section*{Acknowledgements}

This research received funding from the EPSRC project MaDrIgAL (EP/N017536/1). The Titan Xp used for this research was donated by the NVIDIA Corporation.

\bibliography{emnlp2018}
\bibliographystyle{acl_natbib_nourl}

\end{document}